\documentclass[runningheads]{llncs}

\usepackage{eccv}

\usepackage{eccvabbrv}

\usepackage{graphicx}
\usepackage{booktabs}
\usepackage{amsmath}
\usepackage{amssymb}
\usepackage{bm}
\usepackage[numbers]{natbib}
\usepackage{graphicx}
\usepackage{caption}
\usepackage{subcaption}
\usepackage{lipsum}
\usepackage{xcolor}
\usepackage{colortbl}
\usepackage{graphicx}
\usepackage{array}
\usepackage{svg}
\usepackage{float}
\usepackage{booktabs}
\usepackage{multirow}
\usepackage{listings}
\usepackage{amsmath}
\usepackage[table]{xcolor}   

\usepackage[accsupp]{axessibility}  

\usepackage{hyperref}

\usepackage{orcidlink}

\begin{document}

\title{Generalization and Memorization in Rectified Flow} 


\author{Mingxing Rao\orcidlink{0009-0006-6582-4598} \and
Daniel Moyer\thanks{Corresponding author.}\orcidlink{0000-0003-4428-5012}}

\authorrunning{M. Rao and D. Moyer}

\institute{Vanderbilt University, Nashville TN 37235, USA
\\
\email{\{mingxing.rao,daniel.moyer\}@vanderbilt.edu}}

\maketitle

\begin{abstract}
Generative models based on the Flow Matching objective, particularly Rectified Flow, have emerged as a dominant paradigm for efficient, high-fidelity image synthesis. However, while existing research heavily prioritizes generation quality and architectural scaling, the underlying dynamics of how RF models memorize training data remain largely underexplored. In this paper, we systematically investigate the memorization behaviors of RF through the test statistics of Membership Inference Attacks (MIA). We progressively formulate three test statistics, culminating in a complexity-calibrated metric ($T_\text{mc\_cal}$) that successfully decouples intrinsic image spatial complexity from genuine memorization signals. This calibration yields a significant performance surge—boosting attack AUC by up to 15\% and the privacy-critical TPR@1\%FPR metric by up to 45\%—establishing the first non-trivial MIA specifically tailored for RF. Leveraging these refined metrics, we uncover a distinct temporal pattern: under standard uniform temporal training, a model's susceptibility to MIA strictly peaks at the integration midpoint, a phenomenon we justify via the network's forced deviation from linear approximations. Finally, we demonstrate that substituting uniform timestep sampling with a Symmetric Exponential (U-shaped) distribution effectively minimizes exposure to vulnerable intermediate timesteps. Extensive evaluations across three datasets confirm that this temporal regularization suppresses memorization while preserving generative fidelity.
\end{abstract}
   
\section{Introduction} \label{sec:intro}
Flow Matching (FM)~\cite{lipman2022flow} has emerged as a powerful, simulation-free paradigm to overcome the slow sampling bottlenecks of Diffusion Models (DMs)~\cite{ho2020denoising, rombach2022high, song2019generative, song2020score, karras2022elucidating}. Among its variants~\cite{lipman2022flow, albergo2022building, tong2023improving}, Rectified Flow (RF)~\cite{liu2022flow} stands out by connecting data and noise distributions via straight-line trajectories. This linearizes the generation process, enabling highly efficient, few-step sampling. Due to its stability and efficiency, RF underpins contemporary state-of-the-art foundation models such as Stable Diffusion 3~\cite{esser2024scaling}, FLUX~\cite{labs2025flux1kontextflowmatching}, and Ideogram 4~\cite{ideogram-4-2026}. However, while prior work heavily optimizes RF for generation quality~\cite{tong2023improving, pooladian2023multisample} and architectural scaling, its memorization and generalization dynamics are relatively underexplored. 

Conventional indicators of overfitting, such as the standard generalization gap between training and validation losses, fail to accurately capture the complex, time-dependent memorization behaviors inherent to RF (shown in Appendix~\ref{supp:loss_gap}). Even if loss gaps are small, it may be possible to recover the specific training data from model weights, outputs, or behavior/activations during generation. This process is known as model inversion~\cite{rao2025latent, rao2025score, fredrikson2015model}. A specific sub-task of model inversion is classifying samples as drawn from the training set or otherwise using those same model statistics; this is known as membership inference~\cite{carlini2022membership}, and particular methods referred to as membership inference attacks (MIAs).

The ability to perform an MIA against a flow matching model is indicative of memorization; if we can discriminate between training and non-training data using the model itself, even though both training and non-training sets are ostensibly drawn from the same underlying population, there is some training set specific information left in the model.






In this work we construct an MIA for rectified flow. We first establish a baseline test criterion, $T_\text{naive}$, derived from the theoretical Flow Matching (FM) objective. We further formulate a Monte Carlo estimator, $T_\text{mc}$, based on the practical and commonly used Conditional Flow Matching (CFM) objective. We provide detailed motivation and show significant performance improvement for the transition from $T_\text{naive}$ to $T_\text{mc}$.
We then propose a correction/calibration $T_\text{mc\_cal}$ for an image-domain specific phenomenon in generative modelling, spatial complexity biases \cite{nalisnick2018deep, serra2019input}). 

\begin{figure}[t]
  \begin{center}
  \includegraphics[width=0.8\textwidth]{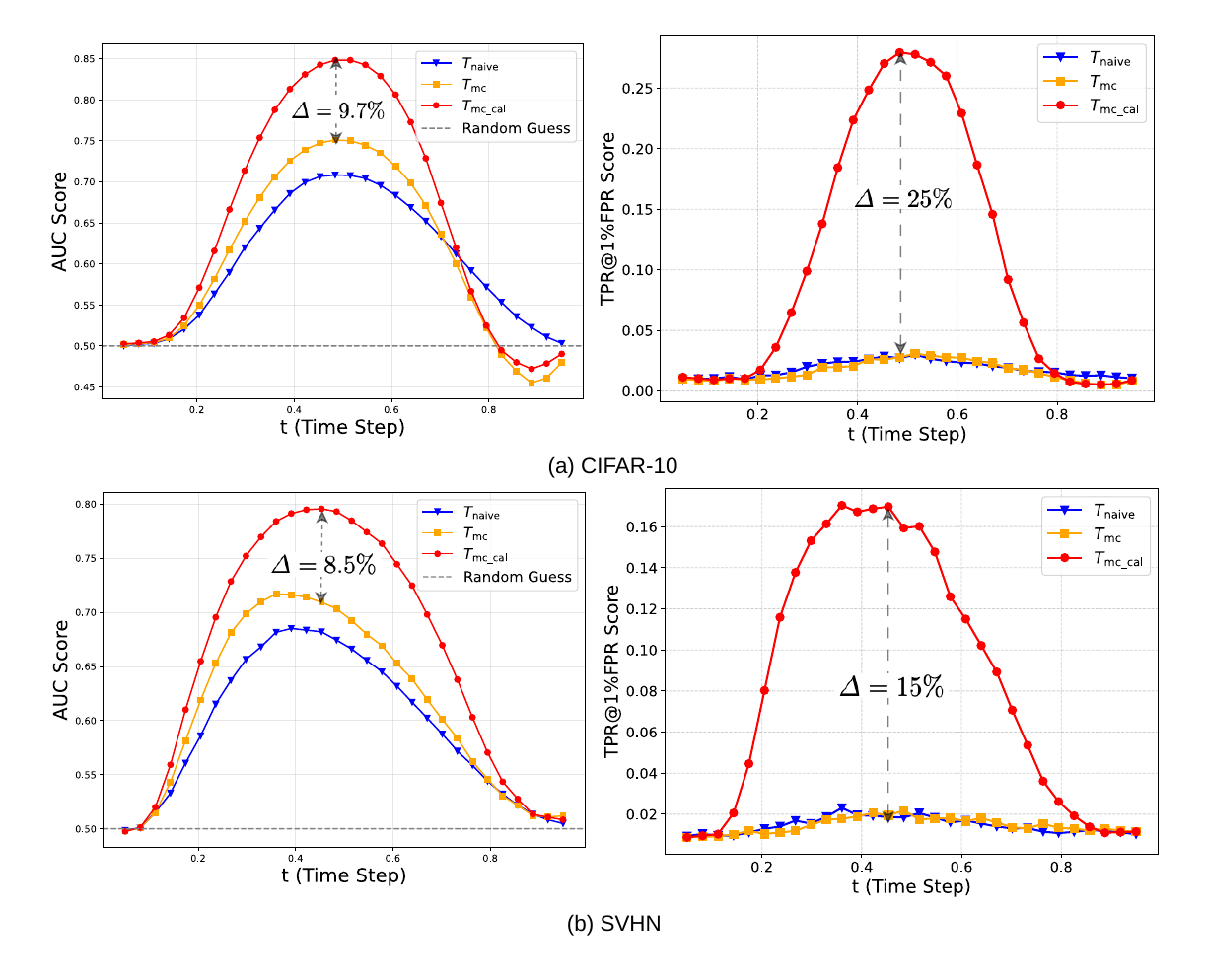}
  \end{center}
  \caption{MIA performance (AUC and TPR@1\%FPR, at \textbf{left} and \textbf{right} respectively) across integration timesteps $t$ on CIFAR-10 and SVHN (\textbf{top} row and \textbf{bottom} row). Our complexity-calibrated statistic ($T_\text{mc\_cal}$) demonstrates a substantial performance surge over the baselines ($T_\text{naive}$ and $T_\text{mc}$).} 
  \label{fig:main_framework}
  
\end{figure}

Utilizing these refined metrics, we uncover a non-trivial temporal pattern in the memorization dynamics of continuous-time models. When employing a standard uniform temporal sampling strategy $t \sim \mathcal{U}(0, 1)$, the model's susceptibility to MIA is heavily concentrated in the intermediate integration stages, and maximal at the midpoint ($t=0.5$) of the flow (Fig \ref{fig:main_framework}. We provide mathematical justification for this phenomenon by analyzing the deviation between the non-linear regressor and a Linear Minimum Mean Square Error (LMMSE) baseline. Specifically, we show that at $t=0.5$, the noisy input state becomes statistically orthogonal to the target velocity. Consequently, the network is forced to bypass trivial linear approximations and exploit its non-linear capacity to memorize sample-specific features, thereby maximizing the memorization problem at this exact juncture.

Finally, we tackle the persistent utility-privacy trade-off that typically exists in classic Differential Privacy (DP) mechanisms in generative modeling (see Section~\ref {subsec:dp}). To mitigate the accumulation of privacy risks without degrading the visual quality of generated samples, we intervene directly in the training dynamics. By substituting the conventional uniform timestep sampling with a Symmetric Exponential (U-shaped) distribution~\cite{lee2024improving}, we deliberately minimize the model's exposure to the highly vulnerable intermediate timesteps. Our theoretical and empirical analyses confirm that this temporal regularization substantially impedes memorization while preserving generative fidelity (Fig.~\ref{fig:sym_exp}).

We conduct all experiments on CIFAR-10~\cite{krizhevsky2009learning}, SVHN~\cite{netzer2011reading}, and TinyImageNet~\cite{deng2009imagenet} datasets. Three datasets consistently justify our proposed MIA test statistics and analysis of memorization.

In summary, our contributions are the following:
\begin{enumerate}
    \item We progressively construct three test statistics ($T_\text{naive}$, $T_\text{mc}$, $T_\text{mc\_cal}$) for membership inference attack and justify each one with our motivation and derivation. We also empirically shows its practical usage.
    \item With the MIA test statistics, we substantiate our claim that, under reasonable assumptions, the expected memorization risk for well-trained rectified flow across the entire continuous timesteps is strictly upper-bounded by the memorization evaluated at the midpoint.
    \item We replace the standard uniform sampling of $t$ with the Symmetric Exponential distribution proposed by Lee et al.~\cite{lee2024improving}, providing a novel and comprehensive justification from the dynamics of model generalization and memorization. 
\end{enumerate}

All data splits, model checkpoints, training, and testing code are released on our GitHub repository \url{https://github.com/mx-ethan-rao/rf_gen_mem.git}.

  
  
  

\section{Background and Motivation} \label{sec:related_work}

\subsection{Membership Inference Attack}
The goal of the Membership Inference Attack (MIA) is to determine whether a specific data record $x$ was sampled from the underlying population distribution to be included in the empirical training set $\mathcal{D}_{train}$ of a target model $f_\theta$. Carlini et al.~\cite{carlini2022membership} formally cast membership inference as a rigorous statistical hypothesis testing problem. A variety of test statistics have been extensively developed for diffusion-based generative paradigms~\cite{rao2025score, duan2023diffusion, kong2023efficient, fu2023probabilistic, zhai2024membership}. However, the research on the diffusion model does not extend to the model trained via flow matching~\cite{lipman2022flow, liu2022flow}. We proceed from first principles (the training objectives) to mathematically derive a novel test statistic inherently tailored to the Rectified Flow~\cite{liu2022flow} framework. Because models suffering from severe overfitting naturally exhibit high susceptibility to MIA, this derived statistic also serves as a quantitative metric to evaluate the dynamics between model generalization and training data memorization~\cite{rao2025latent}.

\subsection{Generalization and Memorization}
The divergence between training and validation loss—formally known in learning theory as the generalization gap—is the most fundamental statistical signature of overfitting. This method is extensively used on memorization analysis of VAE~\cite{van2021memorization}, GAN~\cite{hayes2017logan, yang2022generalization} and Diffusion Model~\cite{bonnaire2025diffusion, kadkhodaie2023generalization, lukoianov2025locality, gu2023memorization, yoon2023diffusion, li2023generalization}. However, a trivial training/val loss gap proves inadequate for capturing the nuanced memorization dynamics inherent to Rectified Flows (RF)~\cite{liu2022flow}, as evidenced by the empirical loss trajectories in Appendix~\ref{supp:loss_gap}. To establish a rigorous analytical framework, we repurpose the test statistic formulation traditionally utilized in Membership Inference Attacks to precisely quantify sample-specific memorization within RF. By using this metric across different timestep $t$, we uncover that memorization is heavily concentrated at the intermediate $t$, whereas the vulnerability remains relatively benign at the temporal extremes (i.e., early and late stages of $t$). We derive a closed-form expression for the exact timestep at which model memorization peaks under the overfitting assumption. We elaborate this time-dependent memorization in Section~\ref{sec:memorization_dynamics}.

\subsection{Differential Privacy}
\label{subsec:dp}
A classic framework to prevent the deep learning model from memorizing individual training samples is Differential privacy~\cite{abadi2016deep}. Abadi et al.~\cite{abadi2016deep} proposed Differential Privacy-Schocastic Gradient Descent (DP-SGD) and momentum accountant that mathematically guarantee that privacy loss accumulates significantly slower over thousands of SGD iterations. However, enforcing differential privacy on generative models inherently introduces a severe privacy-utility (e.g. generative fidelity) trade-off~\cite{dockhorn2022differentially, tramer2020differentially}. This creates a fundamental tension: from a privacy perspective, repeated exposure to an individual sample exacerbates the risk of catastrophic memorization; yet, from a generative modeling standpoint, the network strictly requires extensive exposure to high-quality data to generate good examples. This raises a critical question: \textit{Can we slow down the privacy loss accumulation without sacrificing generative fidelity?} In Section~\ref{subsec:u_shape}, by distorting the sampling distribution (originally, uniform distribution) of the timestep $t$ during the training phase, we empirically show that the model's memorization dynamics are significantly disrupted with the same number of SGD steps. We also provide a theoretical interpretation in section~\ref{sec:memorization_dynamics} for our insight.

\begin{figure}[t]
  \centering
  
  \includegraphics[width=\textwidth]{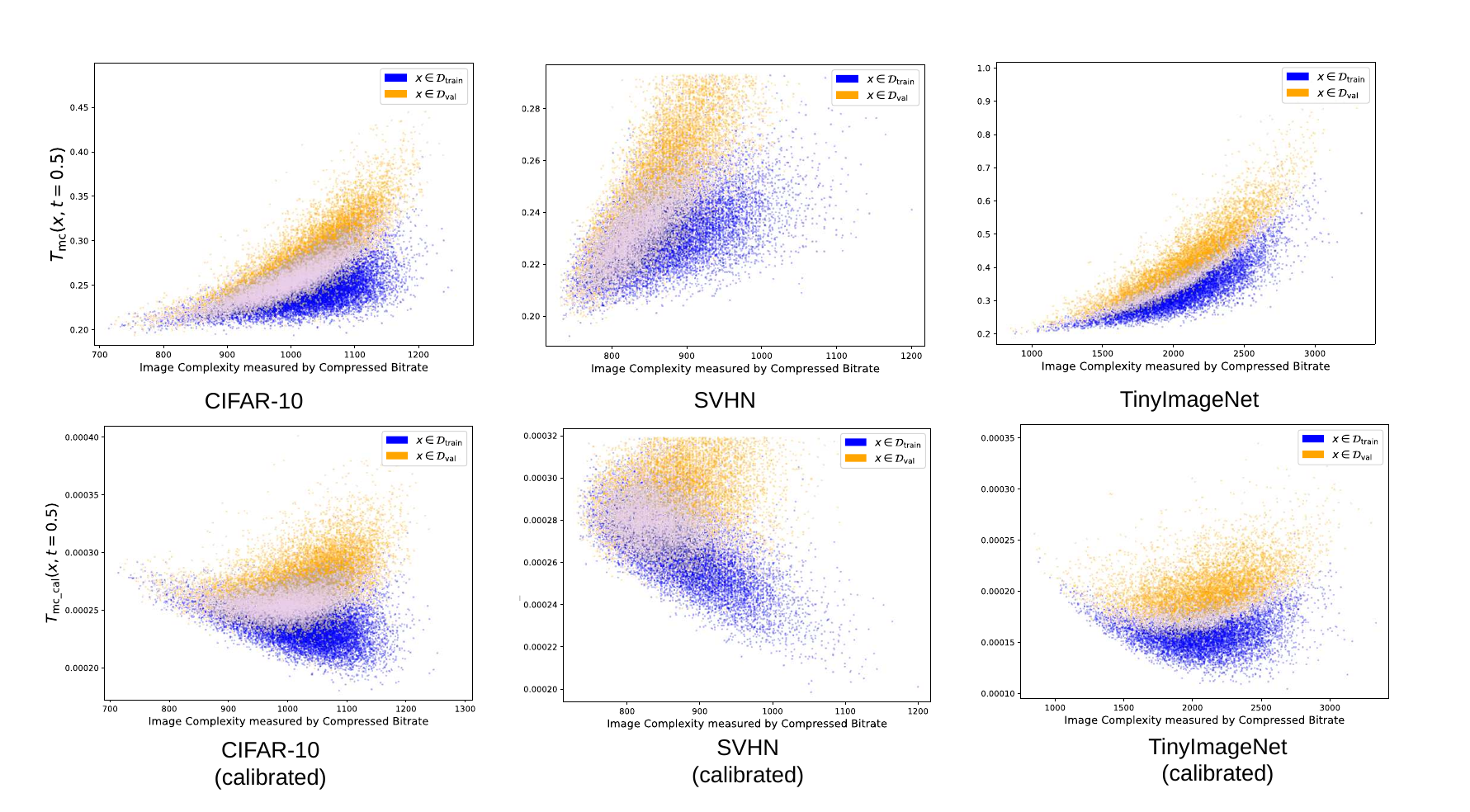}
  
  \caption{\textbf{Top row}: The test statistics $T_\text{mc}$ without calibration. \textbf{Bottom row: } The test statistics $T_\text{mc}$ with calibration. The calibrated version provides more separable points between $\mathcal{D}_\text{train}$ and $\mathcal{D}_\text{val}$}
  
  \label{fig:recon_input_complex_calibrated}
\end{figure}

\subsection{Input Complexity Bias in Generative Models}

Inspired by Out-of-Distribution detection, likelihoods computed from generative models are notoriously susceptible to the inherent complexity of the input data~\cite{nalisnick2018deep, serra2019input, goodier2023likelihood}. Nalisnick et al.~\cite{nalisnick2018deep} first exposed this vulnerability by demonstrating that generative models (Glow~\cite{kingma2018glow}, PixelCNN~\cite{van2016pixel}, and VAE~\cite{kingma2013auto}) often assign higher likelihoods to visually simpler OOD samples (e.g., SVHN) than to the complex in-distribution data they were trained on (e.g., CIFAR-10). Serrà et al.~\cite{serra2019input} formally identified input complexity—often measurable via standard compression bitrates—as the primary confounding factor, revealing that likelihood estimations are heavily dominated by low-level spatial redundancies rather than high-level semantic content. This complexity bias has since been shown to persistently plague modern generative paradigms, including Denoising Diffusion Probabilistic Models~\cite{goodier2023likelihood}. In continuous-time frameworks such as Rectified Flows, the likelihood is computed through the instantaneous change of variables theorem~\cite{grathwohl2018ffjord, lipman2022flow, lipman2024flow}. For held-out dataset $x \in \mathcal{D}_\text{data} \setminus \mathcal{D}_\text{train}$, we confirmed the same phenomenon (See Appendix~\ref{supp:input_bias})on Rectified Flow. The likelihood also served as an important MIA test statistic. This provides insight that the proposed MIA test statistic is heavily biased towards input complexity as well.

\section{MIA Test Statistics for Rectified Flow} \label{sec:methodology}

In this section, we formalize our proposed methodology. We derive our test statistics for membership inference attack from first principles based on training objectives (Section~\ref{subsec:derivation_T}) and introduce a calibration mechanism conditioned on input image complexity (Section~\ref{subsec:calibrate_T}). We justify this calibration strategy and demonstrate its empirical benefits in Section~\ref{subsec:calibrate_T}.



\textbf{Preliminaries:}
The flow matching~\cite{lipman2022flow} framework defines a system of ordinary differential equations (ODEs) which transport particles from a ``base'' distribution $\pi_0$ (usually chosen to be simple, e.g., standard Gaussian noise) to an empirical target distribution $\pi_1$ (e.g., real images). Let $X_0 \sim \pi_0$ and $X_1 \sim \pi_1$ denote the source and target random variables, respectively. Flow matching implicitly defines an interpolent path between these two distributions over a continuous time variable $t \in [0, 1]$ by defining flow fields $v(x,t)$
\begin{equation}
dX_t = v(X_t,t) dt 
\end{equation}
so that $X_{t=0}$ has distribution $\pi_0$, and that $X_{t=1}$ has distribution $\pi_1$. A common choice of field is the Rectified Flow \cite{liu2022flow}, which is defined implicitly by drawing straight lines between pairs of points and calculating the interpolant field:
\begin{equation}
    X_t =  t X_1 + (1-t) X_0
    \label{eq:interpolation}
\end{equation}
While our theoretical analysis extends to other Flow Matching variants (detailed in Appendix~\ref{supp:test_other_fm}), we mainly focus on Rectified Flow, as it is conceptually simple, quite performant, and thereby has become the most popular in contemporary work. 


The time derivative of linear path flows yield a constant vector field pointing directly from $X_0$ to $X_1$, given by $\frac{\mathrm{d} X_t}{\mathrm{d} t} = X_1 - X_0$. During inference, the high level objective of flow matching is to fit a neural network $v_\theta(X _t, t)$ to that vector field condition, fitting an explict function that for any given timestep $t$ and position $x$ can produce the appropriate vector. The original flow matching procedure fit $v_\theta$ by least-squares optimization of the linear path criterion \cite{lipman2022flow}:
\begin{equation}
    \mathcal{L}_{\text{FM}}(\theta) = \mathbb{E}_{t \sim \mathcal{U}(0,1), X_0 \sim \pi_0, X_1 \sim \pi_1} \left[ \left\| (X_1 - X_0) - v_\theta(X_t, t) \right\|^2 \right].
    \label{eq:rf_loss}
\end{equation}
where $\mathcal{U}(0,1)$ denotes the uniform sampling of $t \in [0, 1]$. 

Directly optimizing the original FM objective is computationally intractable due to the integration over a data distribution required to compute the marginal vector field. To circumvent this limitation, Lipman et al.~\cite{lipman2022flow} introduced Conditional Flow Matching (CFM). By conditioning on individual image samples $x_1$, the CFM objective utilizes conditional vector fields that can be efficiently computed and optimized over on a per-sample basis. The CFM objective is
\begin{equation}
    \mathcal{L}_{\text{CFM}}(\theta|x_1) = \mathbb{E}_{t \sim \mathcal{U}(0,1), X_0 \sim \pi_0} \left[ \left\| (X_1 - X_0) - v_\theta(X_t, t) \right\|^2 \mid X_1=x_1\right].
    \label{eq:crf_loss}
\end{equation}
A crucial observation is that $\nabla_{\theta}\mathcal{L}_{FM}(\theta) = \nabla_{\theta}\mathcal{L}_{CFM}(\theta|x_1)$ ~\cite{lipman2022flow, lipman2024flow}, meaning that the CFM objective has the same optima as the original FM objective, and that $\theta$ optimized for one objective have the same relative quality for the other objective.

\subsection{Deriving MIA Test Criteria}
\label{subsec:derivation_T}

\textbf{Naive attack:} Based on the FM objective (Eq. \ref{eq:rf_loss}), we can construct a basic model inversion attack. Given a test image $\bm{x}$, we sample Gaussian noise $\varepsilon$. We write the ``naive'' test criterion for MIA
\begin{equation}
     T_\text{naive}(\bm{x, t})=\left\| \bm{x}-(v_\theta(\bm{x_t}, t)+\varepsilon)\right\|^2 \text{ with } \bm{x_t}=t\bm{x}+(1-t)\varepsilon.
    \label{eq:naive_attack}
\end{equation}
For models minimizing Eq.~\ref{eq:rf_loss}, as the models overfit, we observe $T_\text{naive}(\bm{x}\in\mathcal{D}_{train}) < T_\text{naive}(\bm{x}\in\mathcal{D}_{val})$.
Clearly an ``oracle'' perfect flow would have the same $T$ value in expectation over training and validation datasets, but as the model becomes specialized toward individual datapoints, a classic overfitting pattern, the flow will fail to generalize, and this gap will widen.
As a demonstration, we give the following conceptual experiment: consider a single draw from a Gaussian as the target dataset. The $\mathcal{L}_{FM}$ or $\mathcal{L}_{CFM}$ flow would eventually point towards that draw from all positions, meaning that subsequent draws from the mode of the Gaussian would be pulled away from that mode, even though in expectation they should not move.

The $T_\text{naive}$ criterion includes a chosen/sampled $\varepsilon$. We can take this one step further and derive a Monte Carlo-amortized criterion $T_{mc}$.
Consider a fully overfit model minimizing the CFM loss, i.e., with $\mathcal{L}_{CFM}=0$,
\begin{equation}
    \mathbb{E}_{t \sim \mathcal{U}(0,1), X_0 \sim \pi_0} \left[ \left\| (x_1 - X_0) - v_\theta(X_t, t) \right\|^2_2 \right] = 0.
\end{equation}
which is equivalent to
\begin{equation}
     \mathbb{E}\left[x_1\right] - \mathbb{E}\left[X_0\right] - \mathbb{E}\left[v_\theta(X_t, t)\right] = 0.
\end{equation}
\noindent Since $\pi_0$ is the standard Gaussian distribution, $\mathbb{E}\left[X_0\right]=0$. The sample $x_1$ does not depend on $x_0$, so $\mathbb{E}\left[x_1\right]=x_1$. Thus,
\begin{align}
x_1 - \mathbb{E}\left[v_\theta(X_t, t)\right] =  0.
\end{align}
Or in other words, under the assumption that the CFM loss has been minimized, the empirical distribution samples $x_1$ should approximate $\mathbb{E}\left[v_\theta(X_t, t)\right]$, which gives a way to distinguish the training data from the held-out data.

\textbf{Monte Carlo attack: } Given a test image $x$ and $N$ Monte Carlo (MC) samples $\varepsilon_n \sim \pi_0$, a Monte Carlo version of the naive test statistics is defined

\begin{equation}
     T_\text{mc}(x, t)=\| x-\frac{1}{N}\sum_{n}\left[v_\theta(tx+(1-t)\varepsilon_n, t)\right]\|^2_2
    \label{eq:mc_attack}
\end{equation}

Similar to the naive method, $T_\text{mc}(x\in\mathcal{D}_{train}) < T_\text{mc}(x\in\mathcal{D}_{val})$, if the model is over-fit. 

\subsection{Complexity-Calibrated Test Criteria}
\label{subsec:calibrate_T}

Likelihood estimates for out-of-distribution data derived from deep generative models are biased towards the inherent complexity of the input image~\cite{nalisnick2018deep, serra2019input}. Serr{\`a} et al.~\cite{serra2019input} demonstrated that likelihood scores are positively correlated to the input spatial complexity, often quantified via standard compression bitrates, both within distribution and out-of-distribution. For held-out dataset $x \in \mathcal{D}_\text{data} \setminus \mathcal{D}_\text{train}$, we observe the same phenomena in rectified flow (Appendix ~\ref{supp:input_bias}).

Our proposed test statistics, $T_\text{naive}$ and $T_\text{mc}$, also exhibit these biases. As illustrated in Fig.~\ref{fig:recon_input_complex_calibrated}, visually simple samples systematically yield lower $T_\text{mc}$ values, whereas complex samples produce spuriously high $T_\text{mc}$ values across both the CIFAR-10, SVHN, and TinyImageNet datasets.
It is thus valuable to decouple image complexity from actual memorization signal by calibrating our test statistics. We formalize our complexity-calibrated test statistic, $T_\text{mc\_cal}$:
\begin{equation}
     T_\text{mc\_cal}(x, t)=\frac{T_\text{mc}(x, t)}{C(x)} \text{ with } x_t=tx+(1-t)\varepsilon.
    \label{eq:mc_cal}
\end{equation}
The input complexity, denoted as $C(x)$, is flexibly defined; while its best theoretical definition, Kolmogorov complexity~\cite{li2008introduction}, is not computable in most practical settings, we could also use the byte length of the input image $x$ compressed into a bitstream by any popular compression method. These form an upper bound for the uncomputable Kolmogorov complexity~\cite{li2008introduction}. A parameterized calibration equation is $T_\text{mc}/\text{Complexity}^\beta$, where $\beta$ is the single parameter.

This can be transformed to a log-linear relationship: $log(T_\text{mc}) \sim \beta log(\text{complexity})+c$. We choose JPEG as the compression method, as it results in $\beta=1$ across datasets, removing the need for an additional calibration step to select $\beta$. Additional discussion is relegated to Appendix~\ref{supp:diff_compress_methods}, as other options are available, but those options also result in additional calibration steps. 

By integrating complexity calibration, we achieve substantial empirical gains in distinguishing training members ($x \in \mathcal{D}_\text{train}$) from non-members ($x \in \mathcal{D}_\text{val}$), as detailed in our Experiments section~\ref{subsec:mia_experiments}. Furthermore, we demonstrate in Appendix~\ref{supp:calibrated_T} that calibrating $T_\text{mc}$ yields a significantly more pronounced performance improvement compared to applying an equivalent calibration to the naive baseline, $T_\text{naive}$.

\begin{figure}[t]
  \centering
  
  \includegraphics[width=0.9\textwidth]{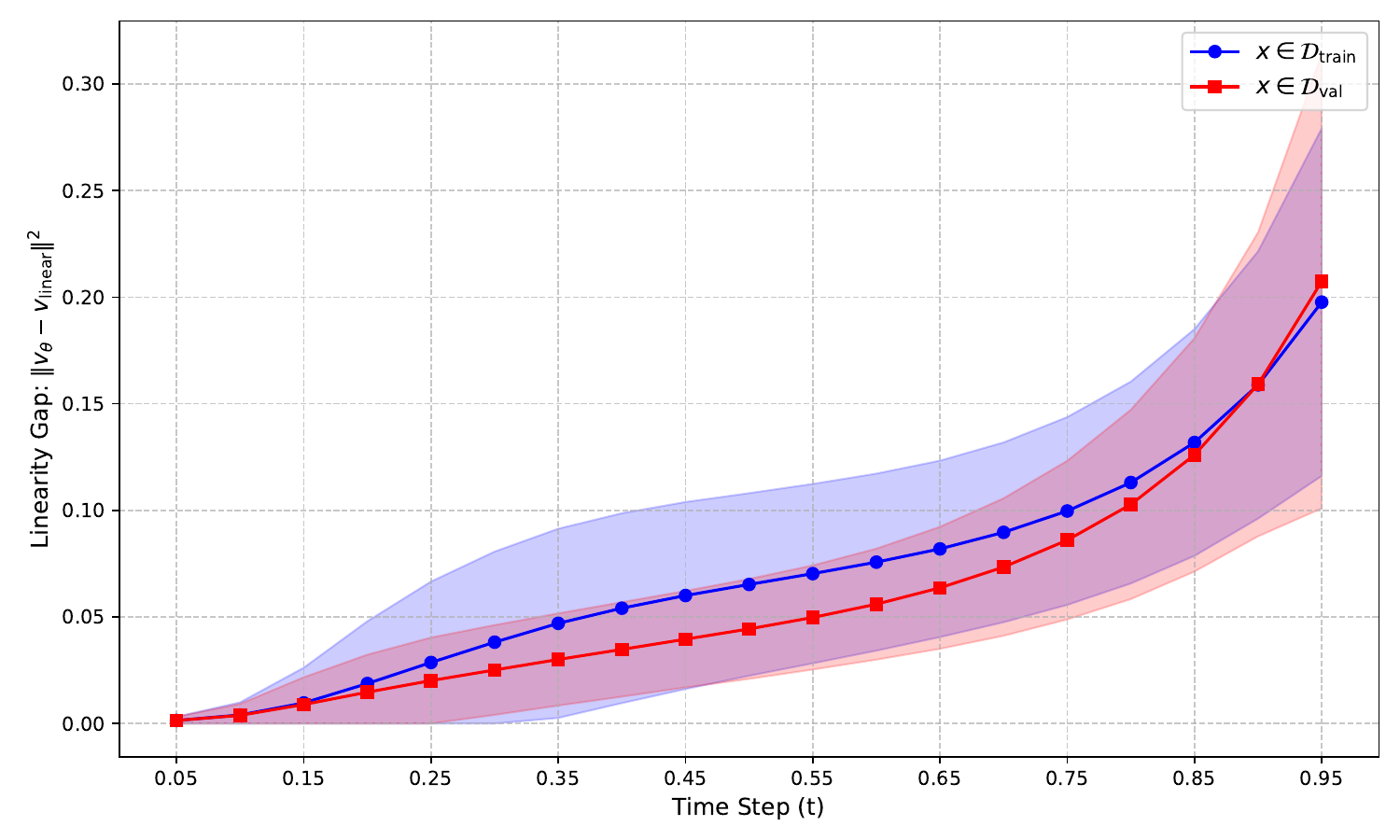}
  
  \caption{The gap between $v_\theta$ and $v_\text{linear}$ for $\mathcal{D}_\text{train}$ and $\mathcal{D}_\text{val}$. Solid lines denote the mean, while the shaded areas indicate $\pm$ standard deviation.}
  
  \label{fig:linearity}
\end{figure}

\section{Memorization Dynamics across Different Timesteps $t$}
\label{sec:memorization_dynamics}

The continuous timestep $t$, as an important input of our proposed test statistics, serves as a critical lens in reflecting overall model memorization. In our empirical evaluations (Fig.~\ref{fig:main_framework}), we surprisingly discover that, under a uniform temporal sampling strategy $t \sim \mathcal{U}(0, 1)$, the model's vulnerabilities to MIA peaks distinctly at the midpoint, around $t=0.5$. This phenomenon is not a spurious empirical artifact; rather, we show that it is intrinsically rooted in the underlying architectural properties of RF. In this section, we theoretically analyze the model memorization (is equivalent to the vulnerability to attacks) across varying timesteps $t$. Moreover, in Section~\ref{subsec:u_shape}, we propose a mitigation strategy designed to enhance robustness against catastrophic overfitting without compromising generative fidelity, as evaluated by the Fréchet Inception Distance (FID).

\subsection{The Linearity Gap between $v_\theta$ and $v_\text{linear}$ across $t$} 

In Rectified Flow, the neural network $v_\theta$ is trained to regress the complex, non-linear vector field. To study the generalization and memorization on rectified flow, we draw inspiration from recent theoretical studies that analyze complex diffusion denoisers through the linear filter, Wiener filter~\cite{yang2023diffusion, li2024understanding, kadkhodaie2023generalization, lukoianov2025locality},
We use the Linear Minimum Mean Square Error (LMMSE) estimator as the linear regressor ($v_\text{linear}$) representing the absolute absence of non-linear feature extraction or memorization. Let the global mean of the data distribution be $\mu_{x_1} = \mathbb{E}[x_1]$ and its global covariance matrix be $\Sigma_{x_1} = \text{Cov}(x_1)$. Recap RF forward process, the state $x_t$ is defined as:
$$
x_t = tx_1 + (1-t)x_0
$$
where $x_0 \sim \mathcal{N}(\mathbf{0}, \mathbf{I})$ is the standard Gaussian noise independent of $x_1$. For non-linear regerssor $v_\theta$, The input information is $x_t$ and the target velocity vector is $v = x_1 - x_0$. To formulate the LMMSE baseline, we formulate it as a constrained optimization problem.

The LMMSE estimator seeks an optimal weight matrix $W^* \in \mathbb{R}^{D \times D}$ and bias vector $b^* \in \mathbb{R}^D$ that minimize the expected squared $\ell_2$-norm of the prediction error:
\begin{equation}
W^*, b^* = \arg\min_{\mathbf{W}, b} \mathbb{E}_{(v, x_t)}\left[ \|v - (Wx_t + b)\|_2^2 \right]
\end{equation}
By solving this constrained objective via the orthogonality principle, we obtain the closed-form global linear predictor:
\begin{equation}
W^* = \Sigma_{vx_t}\Sigma_{x_tx_t}^{-1}, \quad b^* = \mathbb{E}[v] - W^*\mathbb{E}[x_t]
\end{equation}
Substituting the time-dependent moments of the Rectified Flow forward process into this general solution (see Appendix~\ref{supp:derive_llmse} for detailed derivations of the covariance terms), we obtain the closed-form global linear baseline for any input $\bm{x}_t$:
\begin{equation}
v_\text{linear}(x_t) = \mu_{x_1}  + \left( t\Sigma_{x_1} - (1-t)\mathbf{I} \right) \left( t^2\Sigma_{x_1} + (1-t)^2\mathbf{I} \right)^{-1} (x_t - t\mu_{x_1})
\label{eq:v_lmmse}
\end{equation}
We define the \textit{Linearity Gap} $\Delta v$ as the mean squared error between the network's output and the analytical LMMSE prediction:
\begin{equation}
\Delta v(x_t, t) = v_\theta(x_t, t) - v_\text{linear}(x_t)
\end{equation}
A lower $\|\Delta v(x_t, t)\|^2$ suggests a more linear behavior of the vector field $v_\theta(x_t, t)$. By empirically plotting $\|\Delta v\|^2$ across the trajectory $t \in [0, 1]$ in Fig.~\ref{fig:linearity}, it is clear that $\Delta v|_{x \in \mathcal{D}_{val}}$ is lower than $\Delta v|_{x \in \mathcal{D}_{train}}$ and the difference is maximized around intermediate timesteps $t \in [0.45, 0.55]$. This timeframe corresponds to that of peak MIA vulnerability in Fig.\ref{fig:main_framework}. We explain this phenomenon in the next subsection. 

\subsection{Orthogonality between $x_t$ and $v$ at $t=0.5$}
\label{subsec:orthogonality}

$\Sigma_{vx_t}$ is calculated as follows:

\begin{equation}
    \Sigma_{v,x_t} = \mathbb{E}[(v - \mathbb{E}[v])(x_t - \mu)^T]
\end{equation}

Previously, we explicitly compute $\mu_{x_1}$ and $\Sigma_{x_1}$ on $\mathcal{D}_{train}$. For analytical tractability, we assume the data distribution is normalized such that $\mu_{x_1} = \mathbf{0}$ and the covariance $\Sigma_{x_1} = \bm{I}$. This assumption is reasonable as the data is normalized before being sent to the model. Therefore,

\begin{equation}
    \Sigma_{vx_t} = \mathbb{E} \left[ vx_t^\top \right]= \mathbb{E} \left[ (x_1 - x_0)(tx_1 + (1-t)x_0)^\top \right]
\end{equation}
Given the independence of the signal and noise for unseen data ($\mathbb{E}[x_0 x_1^\top] = 0$), this expands to:
\begin{equation}
    \Sigma_{vx_t} = t \mathbb{E}[x_1 x_1^\top] - (1-t) \mathbb{E}[x_0 x_0^\top]
\end{equation}
Since $x_0$ is standard Gaussian ($\Sigma_{x_0} = \bm{I}$):
\begin{equation}
    \Sigma_{x_t, v} = t\bm{I} - (1-t)\bm{I} = (2t - 1)\bm{I}
    \label{eq:covariance}
\end{equation}
\textbf{Observation:} From Eq.~\ref{eq:covariance}, it is evident that at precisely $t=0.5$, the covariance $\Sigma_{x_{0.5}, v} = \bm{0}$. At this timestep, the input $x_{0.5} = 0.5(x_1 + x_0)$ is statistically orthogonal to the target velocity $v = x_1 - x_0$. 

At $t=0.5$ the input provides zero linear information regarding the target and $v_\text{linear}$ degenerates towards mean-prediction (substitute $t=0.5$ to Eq.~\ref{eq:v_lmmse}, $v_\text{linear}(x_{0.5})=\mu_{x_1}=0$). For training samples $x \in \mathcal{D}_\text{train}$, the intermediate timesteps (around $t=0.5$) represent the primary regime where catastrophic memorization occurs. In this heavily mixed state, the model $v_\theta$ can no longer rely on the trivial linear structure of the input. Consequently, during the training phase, the network memorizes the sample-specific structures to minimize the vector field estimation error. At the inference stage, the independence assumption between the target $x_1$ and the model parameters $\theta$ is violated since the network leverages its overfitted weights to condition its prediction on the specific identity of training sample. Conversely, for unseen validation samples $x \in \mathcal{D}_\text{val}$, successful generalization relies entirely on whether these samples share these learned modes~\cite{kadkhodaie2023generalization}. In the absence of such shared structures, the theoretically optimal predictor reduces to the baseline LMMSE estimator, $v_\text{linear}$, causing the network's output to fully collapse to this uninformative state. However, at the temporal boundaries ($t \to 0$ or $t \to 1$), the dynamics shift. Here, $v_\theta$ retains near-perfect visibility of either the pure noise or the original data.

\textbf{Peak Memorization in Rectified Flows: }Assume the integration timestep is uniformly sampled, $t \sim \mathcal{U}(0,1)$, and the target data variable $x_1$ is standardized such that its expectation $\mu_{x_1} = \bm{0}$ and covariance $\Sigma_{x_1} = \bm{I}$. Under these structural conditions, the expected memorization risk across the entire continuous trajectory is strictly upper-bounded by the memorization evaluated at the trajectory midpoint. That is, the maximal memorization occurs at $t=0.5$.

In order to model more general cases where $\Sigma_{x_1}$ departs from $\bm{I}$, we need to generalize Eq.~\ref{eq:covariance} from a zero variance criterion to min variance, i.e., $\min_t \text{Tr}[ t\Sigma_{x_1} + (1-t)\Sigma_{x_0}]$. The optimal $t$ is the point at which the $t$-weighted mean of the eigenvalues of $\Sigma_{x_1}$ and $\Sigma_{x_0}$ is minimal. We demonstrate these cases in Appendix~\ref{supp:toy_dataset}.

We note that a parallel work~\cite{sesmat2026where} independently studies a similar setting, providing analytical evidence for these phenomena using irreducible variance.

\begin{figure}[t]
  \includegraphics[width=\textwidth]{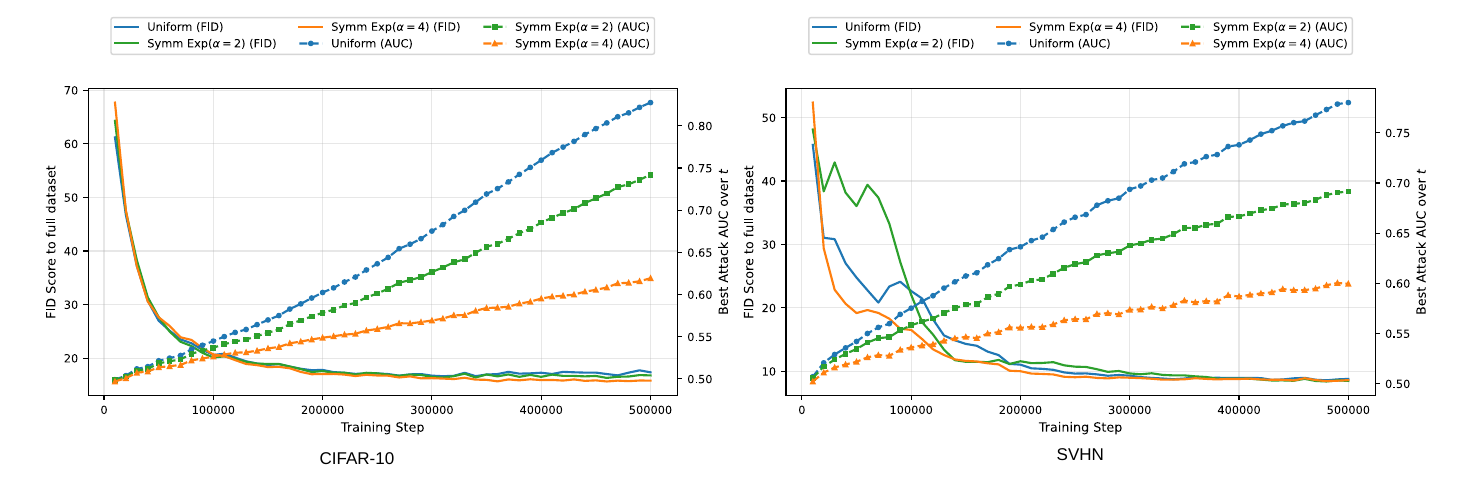}
  \caption{Training dynamics (FID and peak AUC) when employing a Symmetric Exponential distribution for timestep sampling. Increasing the concentration parameter $\alpha$ (shown in order of Blue [none], Green, Orange [most])  effectively decelerates and suppresses model memorization while fully preserving generative fidelity.}
  \label{fig:sym_exp}
\end{figure}

\subsection{U-shape Sampling Distribution for Timesteps $t$}
\label{subsec:u_shape}

To operationalize our insights regarding the memorization dynamics, we replace the conventional uniform sampling $t \sim \mathcal{U}(0, 1)$ with a Symmetric Exponential distribution. This temporal prior explicitly assigns higher probability density to the boundary regions ($t \to 0$ and $t \to 1$) while suppressing the intermediate regime. Formally, for a given concentration parameter $\alpha > 0$, the probability density function (PDF) is defined as:
\begin{equation}
p(t; \alpha) = \frac{\alpha}{2(1 - e^{-\alpha})} \left( e^{-\alpha t} + e^{-\alpha(1-t)} \right), \quad t \in [0, 1]
\end{equation}

\noindent Note that as $\alpha \to 0$, $p(t; \alpha)$ degenerates to the standard uniform distribution. In practice, to ensure numerical stability, we linearly map the sampled $t$ from $[0, 1]$ to a tightly bounded interval $[t_{\min}, t_{\max}]$, where $t_{\min} = 10^{-5}$ and $t_{\max} = 1 - 10^{-5}$. While this technique was initially introduced by Lee et al.~\cite{lee2024improving} based on the empirical observation that the loss of 2-Rectified Flow~\cite{liu2022flow} is high at the boundaries and low in the middle. We justify this with full reasons, that is, from the perspective of model generalization and memorization. For completeness, we show $ \int_{0}^{1} p(t; \alpha) \, dt=1$ in Appendix \ref{supp:proof_normalization}.

While early stopping is a standard regularization technique to mitigate memorization, determining the optimal stopping criterion for generative models remains challenging. Practitioners typically wait for the FID to stabilize; however, as illustrated in Fig.~\ref{fig:sym_exp}, moderate memorization occurs before this convergence. Alternatively, while differential privacy guarantees protection against per-sample memorization, it inherently degrades generative fidelity. Hence, one need for techniques that decelerate the rate of memorization without sacrificing sample quality. In Section~\ref{subsec:exp4ushape}, we empirically demonstrate that our proposed modification successfully achieves this optimal balance.

\begin{figure}[t]
  \centering
  
  \begin{subfigure}{0.49\textwidth}
    \includegraphics[width=\linewidth]{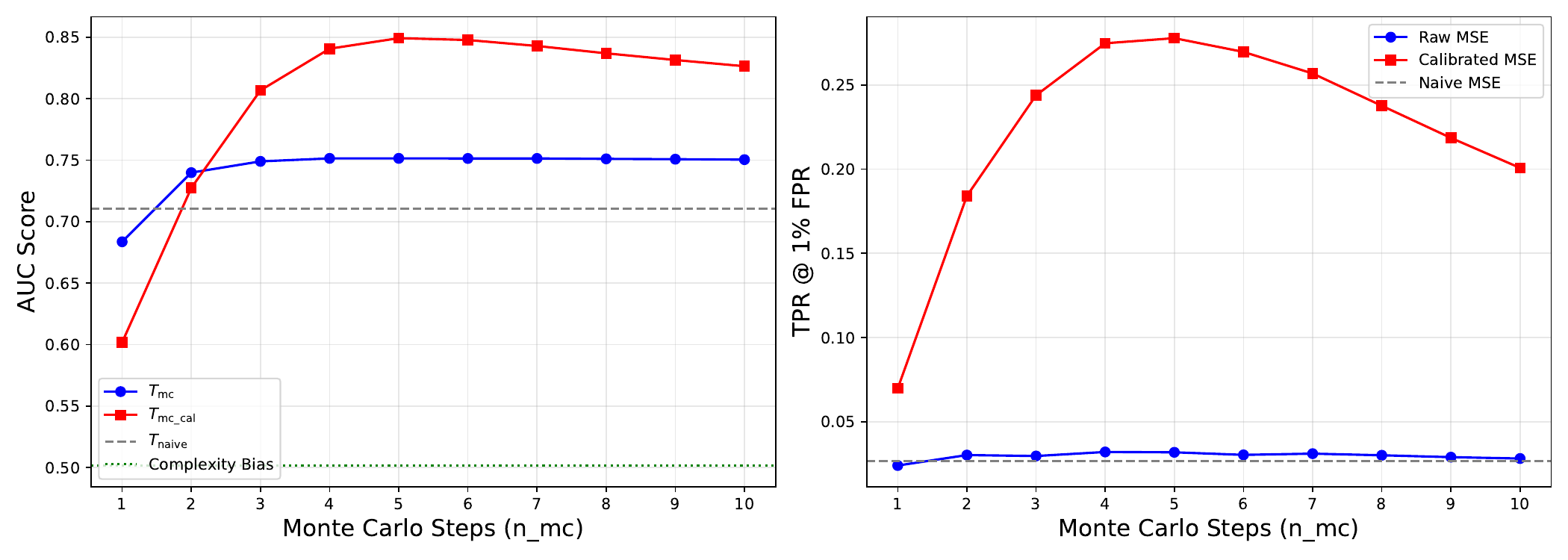}
    \caption{CIFAR-10}
    \label{fig:sub-1}
  \end{subfigure}
  \hfill
  \begin{subfigure}{0.49\textwidth}
    \includegraphics[width=\linewidth]{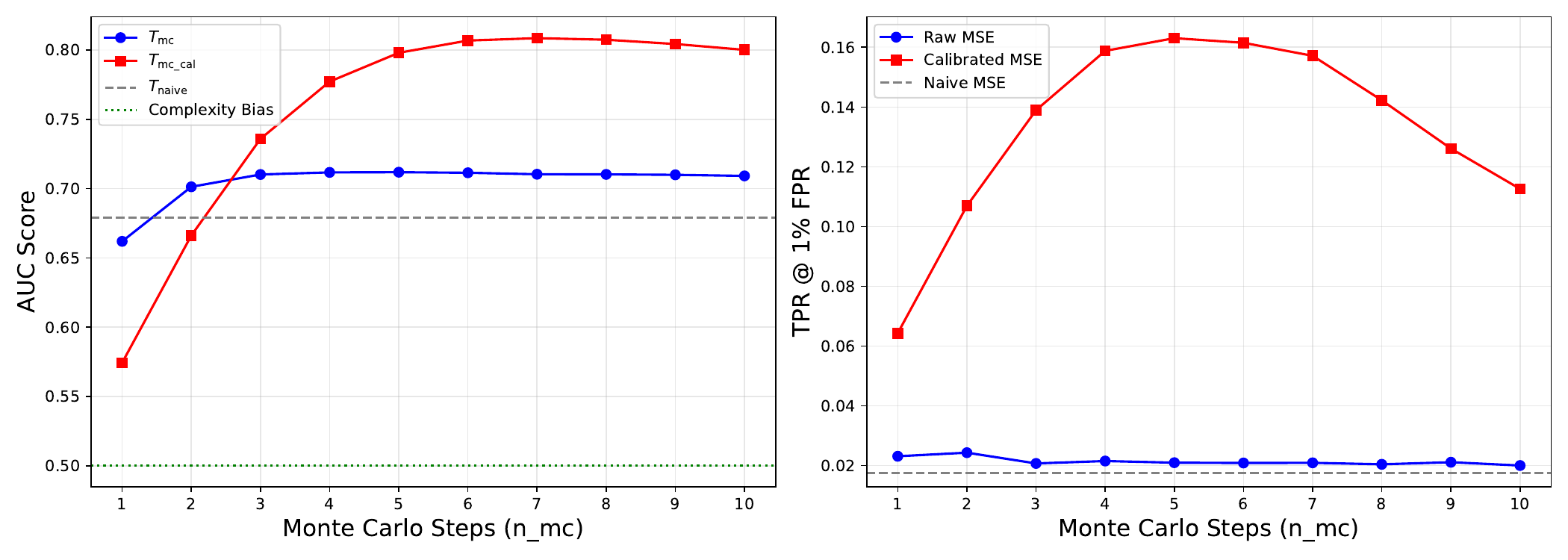}
    \caption{SVHN}
    \label{fig:sub-2}
  \end{subfigure}

  \caption{The dynamic with increasing number of Monte Carlo samples on $T_\text{mc}$ and $T_\text{mc\_cal}$}
  \label{fig:n_mc_sweep}
\end{figure}

\section{Experiments} \label{sec:experiments}

\subsection{Setup}
We tested our findings on three public datasets, CIFAR-10, SVHN, TinyImageNet, MSCOCO with resolution $32^2$, $32^2$, $64^2$, $512^2$ (see Appendix~\ref{supp:dataset_splits} for detailed dataset descriptions and train/validation splits). All experiments are executed on eight NVIDIA A40 GPUs.

\textbf{Evaluation Metrics: }We compute the proposed test statistics for both $\mathcal{D}_{\text{train}}$ and $\mathcal{D}_{\text{val}}$. To quantify memorization, we evaluate the separability between the distributions of $T(x \in \mathcal{D}_{\text{train}})$ and $T(x \in \mathcal{D}_{\text{val}})$. This is measured by the Area Under the ROC Curve (AUC), where we assign binary labels to the samples (e.g., class 1 for $\mathcal{D}_{\text{train}}$ and class 0 for $\mathcal{D}_{\text{val}}$). We report TPR at a 1\% FPR (TPR@1\%FPR) because standard AUC can be misleadingly high even if the attack fails to identify any specific member with high confidence. For computational purposes, the number of queries to compute $T_\text{naive}$, $T_\text{mc}$, and $T_\text{mc\_cal}$ (\#Query). We evaluated the generative image fidelity standard metric, Fréchet Inception Distance.

\textbf{Dataset Configuration:} We follow standard dataset construction protocols for MIA. For CIFAR-10, we evenly split the original training set into 25K/25K samples for $\mathcal{D}_\text{train}$ and $\mathcal{D}_\text{val}$, respectively, and train the RF model for 500K steps on $\mathcal{D}_\text{train}$. Similarly, we construct 20K/20K splits for SVHN (trained for 500K steps), 50K/50K splits for TinyImageNet (trained for 100K steps) and 3K/3K for MSCOCO (trained for 45K steps). To maintain data diversity, we ensure a balanced class distribution across both $\mathcal{D}_\text{train}$ and $\mathcal{D}_\text{val}$. All models are trained using the standard RF scripts from \url{https://github.com/lucidrains/rectified-flow-pytorch.git}.

\begin{table}[t]
\centering
\caption{Correlation analysis between Input Complexity (Compressed Bitrate) and $T_{mc}$ (before calibration). We report Pearson ($\rho$) and Spearman ($r_s$) correlation coefficients for both training and validation sets.}
\label{tab:correlation_analysis}
\setlength{\tabcolsep}{4pt}
\renewcommand{\arraystretch}{1.1}
\small
\begin{tabular}{l cc|cc|cc}
\toprule
\multirow{2}{*}{Split} & \multicolumn{2}{c|}{CIFAR-10} & \multicolumn{2}{c|}{SVHN} & \multicolumn{2}{c}{TinyImageNet64} \\
\cmidrule(lr){2-3} \cmidrule(lr){4-5} \cmidrule(lr){6-7}
& Pearson $\rho$ & Spearman $r_s$ & Pearson $\rho$ & Spearman $r_s$ & Pearson $\rho$ & Spearman $r_s$ \\
\midrule
Train & 0.6030 & 0.5917 & 0.8520 & 0.8699 & 0.5275 & 0.5535 \\
Val   & 0.8318 & 0.8561 & 0.8842 & 0.9060 & 0.7837 & 0.8111 \\
\bottomrule
\end{tabular}
\end{table}

\begin{table*}[!t]
\centering
\caption{MIA Performance comparison across three datasets. All values are reported in percentage (\%). $\Delta$ represents the improvement over the method immediately above.}
\label{tab:mia_performance}
\setlength{\tabcolsep}{4pt} 
\renewcommand{\arraystretch}{1.2}
\scriptsize
\begin{tabular}{lcc|cc|cc}
\toprule
\multirow{2}{*}{Method} &
\multicolumn{2}{c|}{CIFAR-10} &
\multicolumn{2}{c|}{SVHN} &
\multicolumn{2}{c}{TinyImageNet} \\
\cmidrule(lr){2-3}\cmidrule(lr){4-5}\cmidrule(lr){6-7}
& AUC↑ & TPR@1\%FPR↑ & AUC↑ & TPR@1\%FPR↑ & AUC↑ & TPR@1\%FPR↑ \\
\midrule

$T_\text{naive}$  & 70.98 & 2.70 & 68.04 & 1.89 & 72.33 & 4.10 \\
\midrule

$T_\text{mc}$ (\#mc=5) & 75.12 & 3.05 & 70.92 & 1.54 & 76.44 & 5.33 \\
\rowcolor{gray!20}
\quad $\Delta$ (vs. $T_\text{naive}$) & +4.14 & +0.35 & +2.88 & -0.35 & +4.11 & +1.23 \\
\midrule

$T_\text{mc\_cal}$ (\#mc=5)   & \textbf{84.89} & \textbf{27.88} & \textbf{79.43} & \textbf{16.46} & \textbf{92.96} & \textbf{50.03} \\
\rowcolor{gray!20}
\quad $\Delta$ (vs. $T_\text{mc}$) & +9.77 & +24.83 & +8.51 & +14.92 & +16.52 & +44.70 \\

\bottomrule
\end{tabular}
\end{table*}

\begin{table}[t]
  \centering
  \caption{Comparison of membership inference attack (MIA) performance on MSCOCO across different timesteps $t$, for a Rectified Flow model trained in the latent space of a VAE All values are reported in percentage (\%). $\Delta$ represents the improvement over the method immediately above.}
  \label{tab:latent_rf}
  \setlength{\tabcolsep}{3pt}
  \renewcommand{\arraystretch}{1.15}
  \small  
    \begin{tabular}{l|cccccccc}
      \toprule
      $t$
        & 0.10 & 0.20 & 0.30 & 0.40 & 0.50 & 0.60 & 0.70 & 0.80  \\
      \midrule
      $T_\text{naive}$
        & 57.94 & 70.63 & 75.50 & \textbf{77.04} & 76.14 & 73.16 & 68.94 & 48.46  \\
      \midrule
      $T_{\text{mc}}$ (\#mc=5)
        & 59.11 & 73.15 & 80.25 & 85.14 & \textbf{85.33} & 83.45 & 75.63 & 64.37  \\
      \rowcolor{gray!20}
      \quad $\Delta$ (vs.\ $T_\text{naive}$)
       & +1.17 & +2.52 & +4.75 & +8.10 & +9.19 & +10.29 & +6.69 & +15.91  \\
      \bottomrule
    \end{tabular}%
\end{table}

\subsection{Calibration Analysis}
To demonstrate that standard test statistics are highly biased by input image complexity, we plot the likelihood estimation against input complexity in Appendix~\ref{supp:input_bias}, and $T_\text{mc}$ against input complexity in the first row of Fig.~\ref{fig:recon_input_complex_calibrated}. A corresponding correlation coefficient analysis is provided in Table~\ref{tab:correlation_analysis}. Because this inherent bias obscures the true extent of memorization, we decouple the test score from input complexity. As shown in the second row of Fig.~\ref{fig:recon_input_complex_calibrated}, our calibrated test statistic ($T_\text{mc\_cal}$) achieves a significantly cleaner separation between $\mathcal{D}_\text{train}$ and $\mathcal{D}_\text{val}$.

\subsection{MIA Experiments}
\label{subsec:mia_experiments}
We initially evaluate the baseline statistic $T_\text{naive}$ and observe that FM objective-based statistics tend to underestimate memorization. To validate our proposed $T_\text{mc}$, we evaluate its performance by increasing the number of Monte Carlo samples ($N_{mc}$) from 1 to 10. As illustrated in Fig.~\ref{fig:n_mc_sweep}, $T_\text{mc}$ underperforms $T_\text{naive}$ at $N_{mc}=1$, but surpasses it at $N_{mc} \ge 2$ and continues to improve. This behavior suggests that as we sample more noise vectors $\varepsilon$, $\mathbb{E}[\bm{x_0}]$ approaches zero, allowing the velocity network $v_\theta$ to predict the test image $\bm{x_1}$ more directly. We prescribe $N_{mc}=5$ for subsequent experiments to balance computational overhead and AUC performance. Furthermore, because reconstruction error is heavily influenced by image complexity, adopting our calibrated statistic $T_\text{mc\_cal}$ yields a significant performance surge, particularly in TPR@1\%FPR. The final results are summarized in Table~\ref{tab:mia_performance}.

For the dataset with high-resolution images such as MSCOCO, we trained an RF in the latent space of a Variational Autoencoder (VAE). We directly employ the publicly released autoencoder model \texttt{stabilityai/sd-vae-ft-mse}~\cite{stabilityai_sdvae_ft_mse} without further fine-tuning. We swept different $t$ for this experiment and the results are summarized in Table~\ref{tab:latent_rf}. For $T_\text{naive}$, we observe a slight shift towards $t=0$ for optimal $t$ as the images exhibit spatial auto-correlation. For $T_\text{mc}$, optimal $t$ perfectly remains on $0.5$. VAE spaces have regularized covariances, so on average $\Sigma_{x_1}$ will be closer to identity. Notably, $T_\text{mc\_cal}$ is not applicable for latent RF, but we can use the encoder/decoder metric for the calibration. This is described in~\cite{rao2025latent}.

\subsection{Impact of Symmetric Exponential Time Sampling}
\label{subsec:exp4ushape}
To validate the theoretical insights from Section~\ref{subsec:u_shape}, we train additional RF models using a Symmetric Exponential Distribution for time sampling with $\alpha=2$ and $\alpha=4$. Larger values of $\alpha$ concentrate the sampling density toward the boundaries ($t \to 0$ and $t \to 1$). The results are presented in the third column of Fig.~\ref{fig:main_framework}, with supplementary TinyImageNet experiments detailed in Appendix~\ref{supp:tiny_image_exp}. Note that we report the maximum AUC over $t$, which typically occurs near $t=0.5$. The figures clearly demonstrate that introducing this sampling technique significantly decelerates memorization during training, while maintaining or even reducing the FID score. This empirical evidence directly supports our analysis and claims regarding memorization dynamics in Section~\ref{sec:memorization_dynamics}.

 \section{Conclusion and Discussion} \label{sec:conclusion}
In this paper, we present a series of test statistics for MIA on RF and provide a theoretical justification for each refinement. Our analysis of memorization dynamics across timesteps $t$ reveals that memorization is primarily concentrated at the midpoint of ODE trajectory. We justify this observation by establishing a link to the Linear Minimum Mean Square Error (LMMSE) estimator. Leveraging this insight, we adjust the sampling distribution of $t$, a modification that decelerates memorization without compromising generative image fidelity.

\section{Acknowledgments}
\label{sec:Acknowledgments}

This work was supported in part by NSF 2321684 and the ARPA-H ALISS project.
 


%
%
\bibliographystyle{splncs04}
\bibliography{main}

\clearpage
\section{Appendix}

\renewcommand\thesection{\Alph{section}}
\setcounter{section}{0}

\section{Input Complexity Bias in Likelihood Estimation}
\label{supp:input_bias}
\begin{figure}[H]
  \begin{center}
  \includegraphics[width=\textwidth]{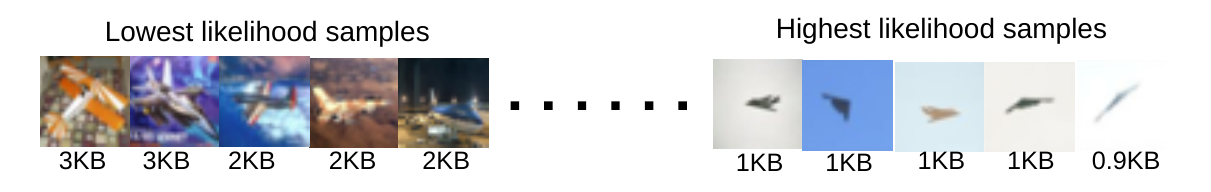}
  \end{center}
  \caption{Examples on $\mathcal{D}_\text{val}$ of highest log-likelihood and lowest log-likelihood in CIFAR-10. The corresponding compressed rate (bytes) is shown for each image.}
  \label{fig:likelihood_complex}
  
\end{figure}

For a trained $v_\theta$ on RF. The evolution of the log-density over time is governed by the instantaneous change of variables theorem, defined by the following ordinary differential equation (ODE):

\begin{equation}
\frac{d \log p(x(t))}{dt} = -\text{Tr}(\nabla_x v_\theta(x(t), t))
\end{equation}
To compute the exact log-likelihood for a given data sample $x_1$, we integrate this ODE backward in time from $t=1$ to the latent state $x_0$ at $t=0$:
\begin{equation}
\log p_1(x_1) = \log p_0(x_0) + \int_{1}^{0} \text{Tr}(\nabla_x v_\theta(x(t), t)) dt
\end{equation}
where $p_0$ represents the standard Gaussian prior distribution $\mathcal{N}(0, I)$. Because computing the exact trace of the Jacobian is computationally expensive for high-dimensional image data, it is approximated using Hutchinson's trace estimator with $K$ independent noise vectors $\epsilon_k \sim \mathcal{N}(0, I)$:
\begin{equation}
\text{Tr}(\nabla_x v_\theta(x(t), t)) \approx \frac{1}{K} \sum_{k=1}^K \epsilon_k^\top \nabla_x v_\theta(x(t), t) \epsilon_k
\end{equation}
We choose $K=2$ for all the experiments. The examples of highest log-likelihood and lowest likelihood is shown in Fig.\ref{fig:likelihood_complex}. It is very clear that the likelihood estimation is biased towards the input complexity. For test images $\mathcal{D}_\text{val}$, less complex images are easily assigned with higher likelihood examples compared to complex images.

\section{Discussion on Different Compression Methods}
\label{supp:diff_compress_methods}

Calibration using $T_\text{mc}/\text{Complexity}^\beta$ ($\beta$=1 across the paper) implies a log-linear relationship: $log(T_\text{mc}) \sim \beta log(\text{complexity}) + c$.
Linear calibration through $\beta$ would have nearly the same performance across other compression methods like WebP or lossless PNG (compressing from bitmap), so long as we included $\beta$ in the division (i.e., $T_\text{mc}/ (\text{Complexity})^\beta$). We shows this in Table~\ref{tab:ablation_compression_method}. However, it is somewhat unreasonable to include a calibration phase in the MIA setting, so instead we choose the most stable of the methods across datasets, JPEG, which has $\beta\approx1$. This is an important topic, but its discussion greatly expands the scope of this paper.

\begin{table}[t]
  \centering
  \caption{Performance comparison of different methods on CIFAR-10 and TinyImageNet.}
  \label{tab:ablation_compression_method}
  \setlength{\tabcolsep}{5pt}
  \small
  \begin{tabular}{l ccc ccc}
    \toprule
    \multirow{2}{*}{Method (oracle)} &
    \multicolumn{3}{c}{CIFAR-10} &
    \multicolumn{3}{c}{TinyImageNet} \\
    \cmidrule(lr){2-4}\cmidrule(lr){5-7}
    & $\beta$ & AUC↑ & TPR↑ & $\beta$& AUC↑ & TPR↑ \\
    \midrule
    $T_\text{mc}$   & - & 75.12 & 3.05 & - & 76.44 & 5.33 \\
    \rowcolor{gray!20}
    $T_\text{mc\_cal}$ (PNG)   & 0.49 & 81.95 & 11.84 & 0.78 & 85.07 & 12.03 \\
    \rowcolor{gray!20}
    $T_\text{mc\_cal}$ (WebP)  & 0.33 & \textbf{84.95} & 22.58  & 0.57& 91.77 & 35.63 \\
    \rowcolor{gray!20}
    $T_\text{mc\_cal}$ (JPEG)  & 1.08 & 83.70 & \textbf{23.35} & 1.14 & \textbf{93.10}  & \textbf{51.11} \\
    \bottomrule
  \end{tabular}
\end{table}

\section{Training/val Loss Gap in Recitified Flow}
\label{supp:loss_gap}

\begin{figure}[H]
  \begin{center}
  \includegraphics[width=\textwidth]{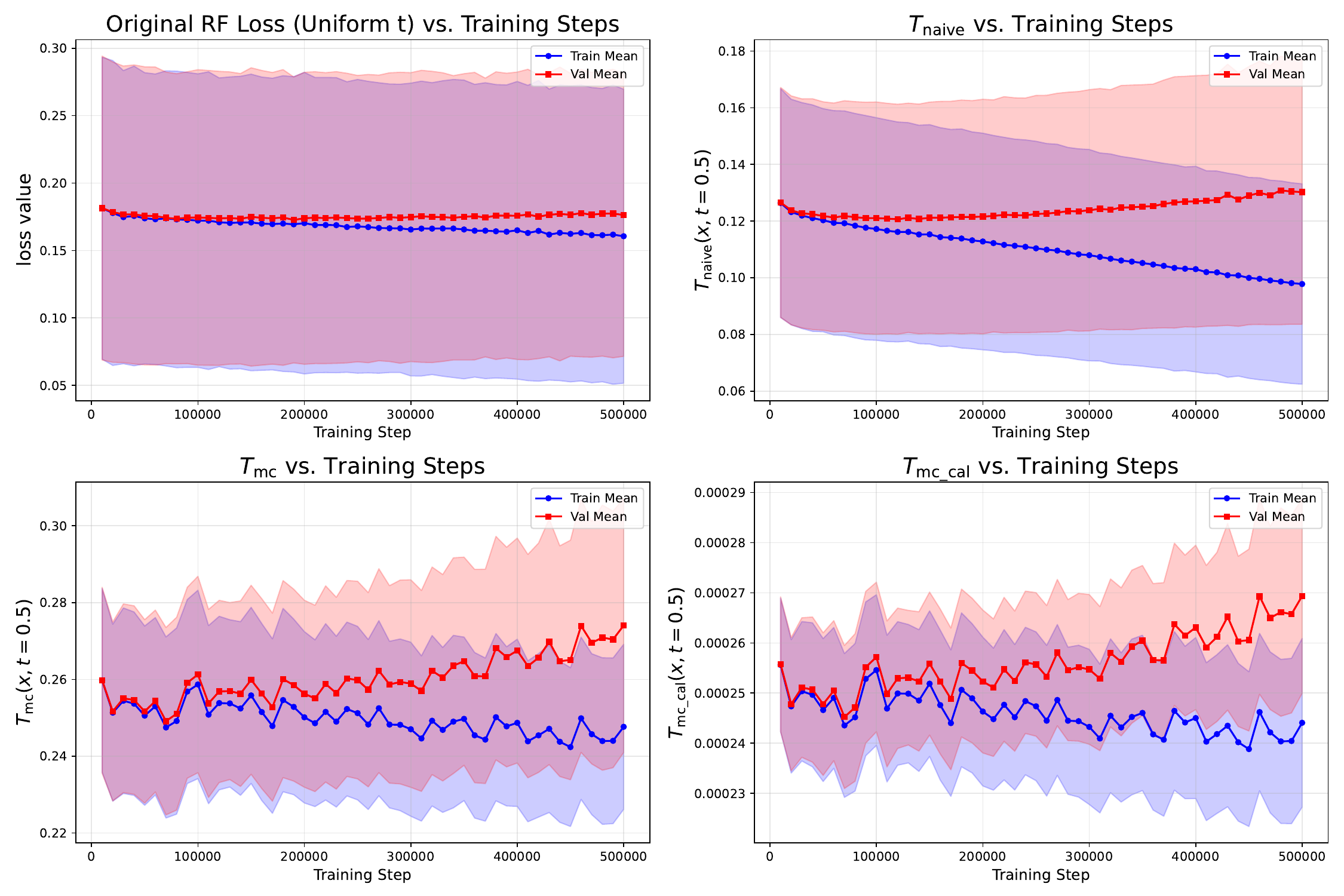}
  \end{center}
  \caption{Top left, top right, bottom left, and bottom right are the curves of $\mathcal{D}_\text{train}$ and $\mathcal{D}_\text{val}$ over CIFAR-10 training steps for original loss, $T_\text{naive}$, $T_\text{mc}$, $T_\text{mc\_cal}$, respectively. The line denotes the mean, and the shadow area denotes the $\pm1 \times \text{std}$. It is clear to see that the test values for $\mathcal{D}_\text{train}$ and $\mathcal{D}_\text{val}$ are progressively more distinguishable and separable.}
  \label{fig:train_val_curve}
  
\end{figure}

As shown in Fig.~\ref{fig:train_val_curve}, the original RF loss exhibits negligible separability between $\mathcal{D}_{\text{train}}$ and $\mathcal{D}_{\text{val}}$, suggested by the near-complete overlap (indicated by the purple region) of their $\pm 1$ standard deviation intervals. This renders it largely ineffective for per-example memorization detection. In contrast, the proposed metrics---$T_{\text{naive}}$, $T_{\text{mc}}$, and $T_{\text{mc\_cal}}$---demonstrate a progressive reduction in this distributional overlap. As the gap between the mean values widens and the overlapping area diminishes, these metrics yield a significantly higher separability for distinguishing members from non-members.

\section{Test Statistics for Optimal Transport conditional FM}
\label{supp:test_other_fm}
This is an example of test statistics for another FM format. For Optimal Transport conditional Vector field, the forward process is~\cite{lipman2022flow}

\begin{equation}
    X_t =  t X_1 + [1-(1-\sigma_{min})t]X_0
\end{equation}
where $\sigma_{min}=\sigma(x_1)$. Then $\frac{\mathrm{d} X_t}{\mathrm{d} t} = X_1 - (1-\sigma_{min})X_0$. Hence, $\mathcal{L}_{\text{CFM}}(\theta|x_1)$ is 
\begin{equation}
    \mathcal{L}_{\text{CFM}}(\theta|x_1) = \mathbb{E}_{t \sim \mathcal{U}(0,1), X_0 \sim \pi_0} \left[ \left\| [X_1 - (1-\sigma_{min})X_0] - v_\theta(X_t, t) \right\|^2 \mid X_1=x_1\right].
    \label{eq:crf_loss}
\end{equation}

\noindent Likewise, consider a fully overfit model minimizing the CFM loss, i.e., with $\mathcal{L}_{CFM}=0$,
\begin{equation}
    \mathbb{E}_{t \sim \mathcal{U}(0,1), X_0 \sim \pi_0} \left[ \left\| [x_1 - (1-\sigma_{min})X_0] - v_\theta(X_t, t) \right\|^2_2 \right] = 0.
\end{equation}
which is equivalent to
\begin{equation}
     \mathbb{E}\left[x_1\right] - (1-\sigma_{min})\mathbb{E}\left[X_0\right] - \mathbb{E}\left[v_\theta(X_t, t)\right] = 0.
\end{equation}

\noindent Since $\pi_0$ is the standard Gaussian distribution, $\mathbb{E}\left[X_0\right]=0$. The sample $x_1$ does not depend on $x_0$, so $\mathbb{E}\left[x_1\right]=x_1$. Thus,
\begin{align}
x_1 - \mathbb{E}\left[v_\theta(X_t, t)\right] =  0.
\label{eq:ot_fm}
\end{align}
Based on Eq.~\ref{eq:ot_fm}, we derive the same $T_\text{mc}$ as that for RF.

\section{Calibration between $T_\text{mc}$ and $T_\text{naive}$}
\label{supp:calibrated_T}
$T_\text{naive\_cal}$ denotes $T_\text{naive}$ with the input complexity calibration. 
\begin{equation}
     T_\text{mc\_cal}(x, t)=\frac{T_\text{naive}}{C(x)} \text{ with } x_t=tx+(1-t)\varepsilon.
    \label{eq:naive_cal}
\end{equation}
From Table~\ref{tab:T_abation}, The calibration yields a substantial performance gain for $T_\text{mc}$, whereas the gain for $T_\text{naive}$ is very limited.

\begin{table}[H]
\centering
\caption{Ablation study on the impact of input complexity calibration across three datasets. $T_{\text{naive\_cal}}$ and $T_{\text{mc\_cal}}$ denote the calibrated versions of $T_{\text{naive}}$ and $T_{\text{mc}}$, respectively, with $\Delta$ indicating the performance difference from their uncalibrated baselines. The results demonstrate that calibration yields substantial performance gains for $T_{\text{mc}}$, whereas the improvements for $T_{\text{naive}}$ are notably limited.}
\label{tab:T_abation}
\setlength{\tabcolsep}{4pt} 
\renewcommand{\arraystretch}{1.2}
\scriptsize
\begin{tabular}{lcc|cc|cc}
\toprule
\multirow{2}{*}{Method} &
\multicolumn{2}{c|}{CIFAR-10} &
\multicolumn{2}{c|}{SVHN} &
\multicolumn{2}{c}{TinyImageNet} \\
\cmidrule(lr){2-3}\cmidrule(lr){4-5}\cmidrule(lr){6-7}
& AUC↑ & TPR@1\%FPR↑ & AUC↑ & TPR@1\%FPR↑ & AUC↑ & TPR@1\%FPR↑ \\
\midrule

$T_\text{naive\_cal}$  & 73.57 & 3.70 & 70.42 & 2.37 & 79.61 & 8.32 \\
\rowcolor{gray!20}
\quad $\Delta$ (vs. $T_\text{naive}$) & +2.59 & +1.0 & +2.38 & +0.48 & +7.28 & +4.22 \\
\midrule

$T_\text{mc\_cal}$ (\#mc=5)   & 84.89 & 27.88 & 79.43 & 16.46 & 92.96 & 50.03 \\
\rowcolor{gray!20}
\quad $\Delta$ (vs. $T_\text{mc}$) & +9.77 & +24.83 & +8.51 & +14.92 & +16.52 & +44.70 \\

\bottomrule
\end{tabular}
\end{table}

\section{Derivation for LMMSE estimator}
\label{supp:derive_llmse}
Let the global mean of the data distribution be $\mu = \mathbb{E}[x_1]$ and its global covariance matrix be $\Sigma = \text{Cov}(x_1)$. Recap RF forward process, the state $x_t$ is defined as:
$$
x_t = tx_1 + (1-t)x_0
$$
where $x_0 \sim \mathcal{N}(\mathbf{0}, \mathbf{I})$ is the standard Gaussian noise independent of $x_1$. For non-linear regerssor $v_\theta$, The input information is $x_t$ and the target velocity vector is $v = x_1 - x_0$ and $\mathbb{E}[v]=\mu$. To formulate the LMMSE baseline, we formulate it as a constrained optimization problem.

The LMMSE estimator seeks an optimal weight matrix $W^* \in \mathbb{R}^{D \times D}$ and bias vector $b^* \in \mathbb{R}^D$ that minimize the expected squared $\ell_2$-norm of the prediction error:
$$
W^*, b^* = \arg\min_{W, b} \mathbb{E}_{(v, x_t)}\left[ \|v - (Wx_t + b)\|_2^2 \right]
$$
By solving this constrained objective via the orthogonality principle—which dictates that the optimal prediction error must be orthogonal to the input space—we obtain the closed-form global linear predictor:
$$W^* = \Sigma_{vx_t}\Sigma_{x_tx_t}^{-1}, \quad b^* = \mathbb{E}[v] - W^*\mathbb{E}[x_t]$$

we first derive the exact analytical moments at any given timestep $t$. The expectation of the state is $\mathbb{E}[x_t] = t\mu$, and the auto-covariance of $x_t$ is given by:
$$
\Sigma_{x_tx_t} = t^2\Sigma + (1-t)^2\mathbf{I}
$$
Similarly, the cross-covariance between the target velocity $v$ and the input state $x_t$ is derived as:
$$
\Sigma_{vx_t} = \mathbb{E}[(v - \mathbb{E}[v])(x_t - \mathbb{E}[x_t])^T] = t\Sigma - (1-t)\mathbf{I}
$$
Using these global statistics, the LMMSE estimator constructs the optimal linear projection for the velocity prediction:
$$
v_{LMMSE}(x_t) = \mathbb{E}[v] + \Sigma_{vx_t} \Sigma_{x_tx_t}^{-1} (x_t - \mathbb{E}[x_t])
$$
Substituting the derived moments into the estimator yields the closed-form global linear baseline for any input $x_t$:
$$
v_{LMMSE}(x_t) = \mu + \left( t\Sigma - (1-t)\mathbf{I} \right) \left( t^2\Sigma + (1-t)^2\mathbf{I} \right)^{-1} (x_t - t\mu)
$$

\section{Proof of Normalization for the Symmetric Exponential Distribution}
\label{supp:proof_normalization}

\begin{proof}
To verify that $p(t; \alpha)$ is a valid probability density function, we must show that its integral over the domain $t \in [0, 1]$ equals exactly $1$:
$$
\int_{0}^{1} p(t; \alpha) \, dt = \frac{\alpha}{2(1 - e^{-\alpha})} \int_{0}^{1} \left( e^{-\alpha t} + e^{-\alpha(1-t)} \right) dt 
$$
By the linearity of integration, we can evaluate the terms within the integrand separately. The definite integral of the forward decay term is evaluated as:
$$
\int_{0}^{1} e^{-\alpha t} \, dt = \left[ -\frac{1}{\alpha} e^{-\alpha t} \right]_{0}^{1} = \frac{1 - e^{-\alpha}}{\alpha} 
$$
For the mirrored decay term, we apply the substitution $u = 1 - t$, which yields $du = -dt$. Adjusting the bounds of integration accordingly, we observe that it evaluates to the exact same result by symmetry:

$$
\int_{0}^{1} e^{-\alpha(1-t)} \, dt = \int_{1}^{0} e^{-\alpha u} (-du) = \int_{0}^{1} e^{-\alpha u} \, du = \frac{1 - e^{-\alpha}}{\alpha} 
$$
Summing the results of these two integrals and multiplying by the leading normalization constant, all terms perfectly cancel:
$$
\int_{0}^{1} p(t; \alpha) \, dt = \frac{\alpha}{2(1 - e^{-\alpha})} \left( \frac{2(1 - e^{-\alpha})}{\alpha} \right) = 1 
$$
Thus, the distribution is strictly normalized.
\end{proof}

\section{Discussion if $\Sigma_{x_1} \neq \bm{I}$}
\label{supp:toy_dataset}
We made Gaussian toy datasets of dimension $d$ with varying condition numbers ($\kappa$) of the covariance matrix $\Sigma_{x_1}[\kappa] = \text{diag}(\kappa, \dots,1)$ with entries linearly space between $\kappa$ and 1, training RF models on each.
As $\kappa$ increases, the peak $t$ shifts towards $t=0$. Reasoning behind this phenomenon starts in Eq. 18 of the main text, where we consider mixture distributions $tx_1 + (1-t)x_0$ and their covariance. For the $\Sigma_{x_1}[\kappa] \neq I$ assumption, we need to generalize Eq. 18 from a zero variance criterion to min variance, i.e., $\min_t \text{Tr}[ t\Sigma_{x_1} + (1-t)\Sigma_{x_0}]$. The optimal $t$ is the point at which the $t$-weighted mean of the eigenvalues of $\Sigma_{x_1}$ and $\Sigma_{x_0}$ is minimal. As $\kappa$ becomes large, the optimal $t$ necessarily grows small. The results are summarized in Table~\ref{tab:condition_peak_t}.

We hypothesize that it is unlikely to see very high $\kappa$ in large natural image datasets without pathological problems (this implies the image pixels are highly correlated), but other fields may see these phenomena regularly. Plus, images with high resolution are usually trained on the latent space of VAE. The latent code is naturally regularized, and its covariance matrix is close to the identity matrix.

\begin{table}[t]
  \centering
  \caption{Results of toy datasets of dimension $d$ with varying condition numbers ($\kappa$) of the covariance matrix $\Sigma_{x_1}[\kappa] = \text{diag}(\kappa, \dots,1)$ with entries linearly space between $\kappa$ and 1. }
  \label{tab:condition_peak_t}
  \setlength{\tabcolsep}{4.5pt} 
  \small 
  \begin{tabular}{l cccccccc}
    \toprule
    Condition $\kappa$ & 1.0 ($\Sigma=I$) & 5.0 & 10.0 & 20.0 & 30.0 & 50.0 & 70.0 & 100.0 \\
    \midrule
    Peak $t$ & 0.50 & 0.41 & 0.32 & 0.29 & 0.25 & 0.24 & 0.21 & 0.19 \\
    \bottomrule
  \end{tabular}
\end{table}

\section{TinyImageNet Experiments}
\label{supp:tiny_image_exp}

\begin{figure}[t]
  \begin{center}
  \includegraphics[width=0.9\textwidth]{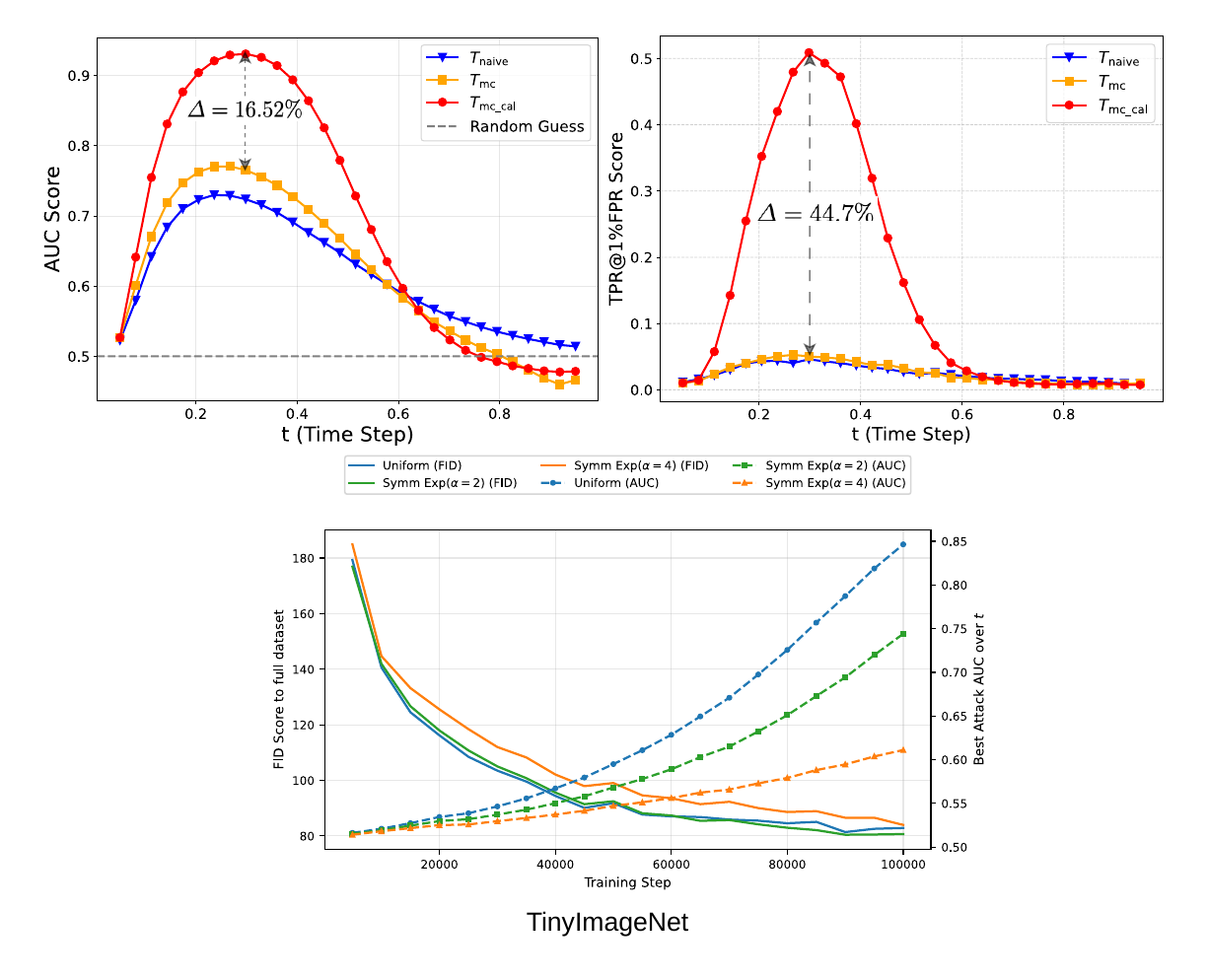}
  \end{center}
  \caption{The MIA performance and training dynamics for TinyImageNet dataset. The performance gain is consistent with that in CIFAR-10 and SVHN. Notably, TPR@1\%FPR increases by 44.7\% after calibration.}
  \label{fig:tinyimnet}
  
\end{figure}

Fig.~\ref{fig:tinyimnet} shows the plots for TinyImageNet. The time of peak memorization is off the midpoint as the assumption ``\textit{the target data variable $x_1$ is standardized such that its expectation $\mu_{x_1} = \bm{0}$ and covariance $\Sigma_{x_1} = I$}'' is broken here. The performance gain is consistent with that in CIFAR-10 and SVHN. Notably, TPR@1\%FPR increases by 44.7\% after calibration.

\section{Dataset and Splits}
\label{supp:dataset_splits}

The UNet architecture and dataset splits are summarized in Table~\ref{tab:unet_architecture} and Table~\ref{tab:dataset_splits}.

\begin{table}[H]
\centering
\caption{UNet Architecture used in our experiments.}
\label{tab:unet_architecture}
\resizebox{\linewidth}{!}{%
\begin{tabular}{l c c c}
\toprule
\textbf{Hyperparameter} & \textbf{UNet 1} & \textbf{UNet 2} & \textbf{UNet 3} \\
\midrule
Base Channels (\texttt{dim}) & 128 & 192 & 128 \\
Channel Multipliers (\texttt{dim\_mults}) & (1, 2, 2, 2) & (1, 2, 3, 4) & (1, 2, 4, 8) \\
Residual Streams (\texttt{num\_residual\_streams}) & 2 & 3 & 2 \\
Input Channels (\texttt{channels}) & 3 & 3 & 4 \\
Attention Head Dim. (\texttt{attn\_dim\_head}) & 32 & 64 & 32 \\
Attention Heads (\texttt{attn\_heads}) & 4 & 4 & 4 \\
Full Attention (\texttt{full\_attn}) & Inner-most & (False, False, True, True) & Inner-most \\
Flash Attention (\texttt{flash\_attn}) & True & True & True \\
Dropout (\texttt{dropout}) & 0.1 & 0.1 & 0.1 \\
\bottomrule
\end{tabular}%
}
\end{table}

\begin{table}[H]
\centering
\caption{Datasets, network architectures, and training configurations used for our experiments.}
\label{tab:dataset_splits}
\resizebox{\textwidth}{!}{%
\begin{tabular}{l c c c c c}
\toprule
\textbf{Dataset} & \textbf{Splits} & \textbf{Resolution} & \textbf{\# Samples / Class} & \textbf{Training Steps} & \textbf{UNet Arch.} \\
\midrule
CIFAR-10 & 25k/25k & 32 & 2500 & 500k & UNet 1 \\
SVHN & 20k/20k & 32 & 2000 & 500k & UNet 1 \\
TinyImageNet & 50k/50k & 64 & 50 & 100k & UNet 2 \\
MSCOCO & 3k/3k & 512 & - & 45K & UNet 3 \\
\bottomrule
\end{tabular}%
}
\end{table}

\end{document}